\documentclass[letterpaper, 10 pt, conference]{ieeeconf}  

\IEEEoverridecommandlockouts                              

\usepackage{cite}
\usepackage{amsmath,amssymb,amsfonts}
\usepackage{algorithmic}
\usepackage{graphicx}
\usepackage{textcomp}
\usepackage{xcolor}
\usepackage{color, colortbl}
\usepackage{todonotes}
\usepackage{multirow}
\usepackage{siunitx}
\usepackage{subcaption}
\usepackage{float}
\usepackage{stfloats}
\usepackage{hyperref}
\usepackage{fancyhdr}
\usepackage{booktabs}
\usepackage{microtype}

\hypersetup{
    colorlinks=true,
    linkcolor=black,
    filecolor=magenta,      
    urlcolor=cyan,
    pdftitle={S2S-Net},
    pdfpagemode=FullScreen,
    }

\title{\LARGE \bf
S2S-Net: Addressing the Domain Gap of Heterogeneous Sensor Systems in LiDAR-Based Collective Perception
}

\author{Sven Teufel$^{1}$, Jörg Gamerdinger$^{1}$ and Oliver Bringmann$^{1}$
\thanks{$^{1}$University of Tübingen, Faculty of Science, Department of Computer Science, Embedded Systems {\tt\small \{sven.teufel, joerg.gamerdinger, oliver.bringmann\} @uni-tuebingen.de}}%
}

\begin{document}

\maketitle
\thispagestyle{empty}
\pagestyle{empty}

\begin{abstract}
Collective Perception (CP) has emerged as a promising approach to overcome the limitations of individual perception in the context of autonomous driving. Various approaches have been proposed to realize collective perception; however, the Sensor2Sensor domain gap that arises from the utilization of different sensor systems in Connected and Automated Vehicles (CAVs) remains mostly unaddressed. This is primarily due to the paucity of datasets containing heterogeneous sensor setups among the CAVs. The recently released SCOPE datasets address this issue by providing data from three different LiDAR sensors for each CAV. This study is the first to  address the Sensor2Sensor domain gap in vehicle-to-vehicle (V2V) collective perception. First, we present our sensor-domain robust architecture S2S-Net. Then an in-depth analysis of the Sensor2Sensor domain adaptation capabilities of state-of-the-art CP methods and S2S-Net is conducted on the SCOPE dataset. This study shows that, all evaluated state-of-the-art mehtods for collective perception highly suffer from the Sensor2Sensor domain gap, while S2S-Net demonstrates the capability to maintain very high performance in unseen sensor domains and outperforms the evaluated state-of-the-art methods by up to 44 percentage points.
\end{abstract}

\section{Introduction}
\label{sec:intro}
In order to operate safe in complex and challenging driving scenarios, automated vehicles (AVs) need a comprehensive perception of their surroundings. In real-world traffic this is not possible in general when using only the local sensor data due to line of sight constraints, limited sensing ranges~\cite{volk_environment-aware_2019} or adverse weather conditions~\cite{teufel2022simulating, teufel2023enhancing}. To overcome these limitations, collective perception (CP) is a promising approach, where information about the environment is shared between Connected and Automated Vehicles (CAVs). The major challenge in CP is the fusion of the collectively shared information with the locally perceived information. CP can be divided into three categories, early, intermediate and late fusion. For late fusion only object states of detected objects are shared which is bandwidth efficient and not affected by sensor domain gaps; however, due to the information loss the performance is usually lower compared to the other fusion methods. For the early fusion raw sensor data such as LiDAR point clouds are exchanged, this poses no information loss and achieves high performance; however, the bandwidth requirement is very high. The intermediate fusion, where neural network features are exchanged, is both bandwidth efficient and achieves very high performance; however, it suffers from feature domain gaps, if not all CAVs run the same models \cite{xu2023bridging}. But not only the different models in the intermediate fusion can cause domain gaps, also the utilization of different sensors among the CAVs can cause a sensor domain gap. Especially for LiDAR sensors, where a high variety of resolutions and Field of Views (FoVs) is available, the domain gap between sensors can become very large. Both the early fusion and the intermediate fusion are prone to this change in the sensor domain. 

In real-world applications manufacturers will deploy a manifold of different sensor systems, while current state-of-the-art methods for CP only consider homogeneous sensor setups among all CAVs. This is mainly caused by previous CP datasets containing data from only one type of sensor; however, the recently released SCOPE \cite{SCOPE-Dataset} dataset contains data from three different LiDAR sensors for each CAV, allowing for an extensive study of the Sensor2Sensor domain gap in CP.

Therefore, we investigate the Sensor2Sensor domain gap in CP and furthermore present a sensor-domain robust fusion architecture called \textbf{S2S-NET}.

\ \\
\ \\[-0.3cm]
Our main contributions are:\\[-0.3cm]
\begin{itemize}
    \item We provide the first study on Sensors2Sensor domain adaptation in LiDAR-based V2V Collective Perception\\[-0.3cm]
    \item We present S2S-Net, a domain robust fusion architecture which fuses the shared sparse voxel grids from various CAVs.\\[-0.3cm]
    \item S2S-Net outperforms all other state-of-the-art methods in domain adaptation on the SCOPE 3D object detection benchmark.\\[-0.3cm]
\end{itemize}
\ \\
\ \\[-0.7cm]

In Section~\ref{rel_work} we introduce related work to our approach. Afterwards, in Section \ref{sec:s2snet}, we present S2S-Net, a domain robust fusion architecture for LiDAR-based collective perception. Section~\ref{sec:eval} describes the conducted experiments and the evaluation. The results are provided in Sec.~\ref{sec:results}. Finally, we give a conclusion and outlook on future research.

\section{Related Work}
\label{rel_work}

\subsection{Datasets}
In order to study the domain gap arising from different sensor domains in CP, datasets containing data from different sensor types are required. 
Axmann et al.~\cite{lucoop} published the LUCOOP real-world V2V dataset. The dataset consists of recordings from three CAVs equipped with different LiDAR sensors. The first CAV is equipped with two LiDAR sensors. One of these is a Hesai Pandar 64-layer LiDAR mounted horizontally, and the other is a Velodyne VLP16 16-layer LiDAR mounted vertically at the back of the CAV. The second CAV is equipped with a Hesai Pandar-XT32 32-layer LiDAR and the last CAV with a Velodyne VLP16 LiDAR. All LiDARs recorded at a frequency of 10 Hz. The number of frames recorded for each CAV is as follows: 15,000 for CAV1 and CAV2, and 7,000 for CAV3. It should be noted that the three CAVs drove the same inner-city round course as a convoy multiple times for the purpose of dataset recording. This results in the relative positioning being always similar and the scenario diversity being very low. The consequence of this is that the dataset is not suitable as a benchmark.

V2X-Real, as proposed by Xiang et al.~\cite{xiang2024v2x}, is a real-world V2X dataset. For recording two CAVs are equipped with a Robosense Ruby Plus 128 channel \SI{360}{\degree} LiDAR with \SI{200}{\metre} range and two RSUs are equipped with Ouster OS1-128/64 channel \SI{360}{\degree} LiDAR sensors with \SI{40}{\metre} range. Both, CAV and RSU are additionally equipped with 2-4 1920$\times$1080\,px RGB cameras. In total, the dataset consists of 33k frames containing objects from 10 different classes, including various vehicles, pedestrians and cyclists. However, the dataset only features two CAVs which are equipped with the same 128-layer LiDAR sensor. Consequently, the dataset is not suited for research on sensor domain adaptation in V2V CP. Furthermore, the scenario diversity is severely constrained due to the utilization of a single intersection for data acquisition.

The TUMTraf V2X Collective Perception real-world dataset~\cite{zimmer2024tumtraf}, which was recorded at a large intersection using one RSU and one CAV, consists of 1000 frames. One Ouster LiDAR OS1-64 64-layer \SI{360}{\degree} LiDAR with \SI{120}{\metre} range and four 1920$\times$1200\,px RGB cameras are mounted at a sign bridge as RSU. The CAV is equipped with a Robosense RS-LiDAR-32 \SI{360}{\degree} LiDAR with 32-layers and \SI{200}{\metre} range. Additionally, the CAV is equipped with a 1920$\times$1200\,px RGB camera. However, due to the limited number of frames and the absence of additional CAVs, the dataset is not suitable for Sensor2Sensor domain adaptation in V2V collective perception.
    
In 2022, Yu et al.~\cite{yu2022dair} presented the DAIR-V2X dataset which is a family of different subdatasets in which the DAIR-V2X-C is for vehicle-infrastructure CP. It was the first real-world large-scale collective perception dataset. For recording 28 different intersections were equipped with two 300-layer \SI{100}{\degree} LiDAR sensors with \SI{280}{\metre} sensing range and two 1920$\times$1080\,px RGB cameras recording with \SI{10}{\hertz}. In addition, one CAV is equipped with a 40-layer \SI{360}{\degree} LiDAR and a 1920$\times$1080\,px RGB camera. The dataset consists of 13k frames and features varying object classes including pedestrians and cyclists. However, as DAIR-V2X-C includes only a single vehicle, it is not suitable for evaluating Sensor2Sensor domain adaptation in V2V collective perception.
    
Gamerdinger et al.~\cite{SCOPE-Dataset} published the synthetic SCOPE dataset, which is generated using the CARLA simulator~\cite{carla}. The dataset features a wide range of environments, including urban intersections, rural roads and motorways, and consists of 44 scenarios with a total of 17,600 frames depicting different object classes, such as cars, vans, (motor-)cyclists and pedestrians. The number of CAVs varies between the scenarios, ranging from 3 to 21, as a reasonable traffic density a CAV rate of about \SI{50}{\percent} is chosen. These CAVs are equipped with 5 cameras with 1920$\times$1080\,px and three different LiDAR sensors, one Velodyne HDL64 64-layer \SI{360}{\degree}, one Velodyne VLP32 32-layer \SI{360}{\degree} and a Blickfeld CUBE solid-state LiDAR with 52 lines and a FoV of \SI{70}{\degree}$\times$\SI{30}{\degree}. The CARLA LiDAR model lacks realistic physical properties, such as beam divergence. For SCOPE, an improved LiDAR sensor model~\cite{rosenberger2020sequential} which includes physical properties is used. In addition to CAVs, RSUs are employed by SCOPE in suitable scenarios, such as urban intersections. The RSUs are equipped with the same sensors as the CAVs. The SCOPE dataset is highly suitable for investigating sensor-to-sensor domain adaptation due to the enhanced sensor models, three different LiDAR sensors per sensor unit, the extensive scenario diversity and the wide range of different road users.

\subsection{Domain Adaptation in Collective Perception}
To the current date the sensor domain gap in CP is mostly unstudied; however, recently Wang et al.~\cite{wang2025v2x} published a framework called V2X-DGPE, where they address the domain gap in vehicle-to-infrastructure (V2I) based perception.
They use the DAIR V2X \cite{yu2022dair} dataset for their evaluation.

V2X-DGPE utilizes a shared feature extraction backbone for both the vehicle sensor data as well as the infrastructure data. Since the shared backbone is trained on both sensor domains, no evaluation of the domain adaptation capabilities to unseen sensor domains was conducted, which poses the biggest challenge for the future deployment of CAVs, since the model cannot be trained on all possible sensor domains.

Zhi et al.~\cite{zhi2025cross} also addressed the V2I sensor domain gap on the DAIR V2X dataset by training their DCGNN framework separately on the vehicle data and infrastructure data. During testing, data from both domains was used as input. Their results on the DAIR V2X dataset show a huge performance drop of the infrastructure model compared to the vehicle model evaluated on the cooperative data. Their results highlight the strong impact of the sensor domain gap on the performance of CP. However, since baseline results of vehicle only and infrastructure only perception are missing, it is not possible to quantify the impact. Furthermore, they evaluated their method on a self-generated CARLA \cite{carla} based dataset with 128 layer and 32 layer LiDAR sensors, where their approach showed much better performance; however, since essential information about the dataset are not provided and also baselines of local detection are missing, it is unclear what causes the improved performance.

\section{S2S-Net}
\label{sec:s2snet}
\begin{figure*}
    \centering
    \includegraphics[trim= 1.1cm 3cm 1.9cm 3.5cm, clip, width=\textwidth]{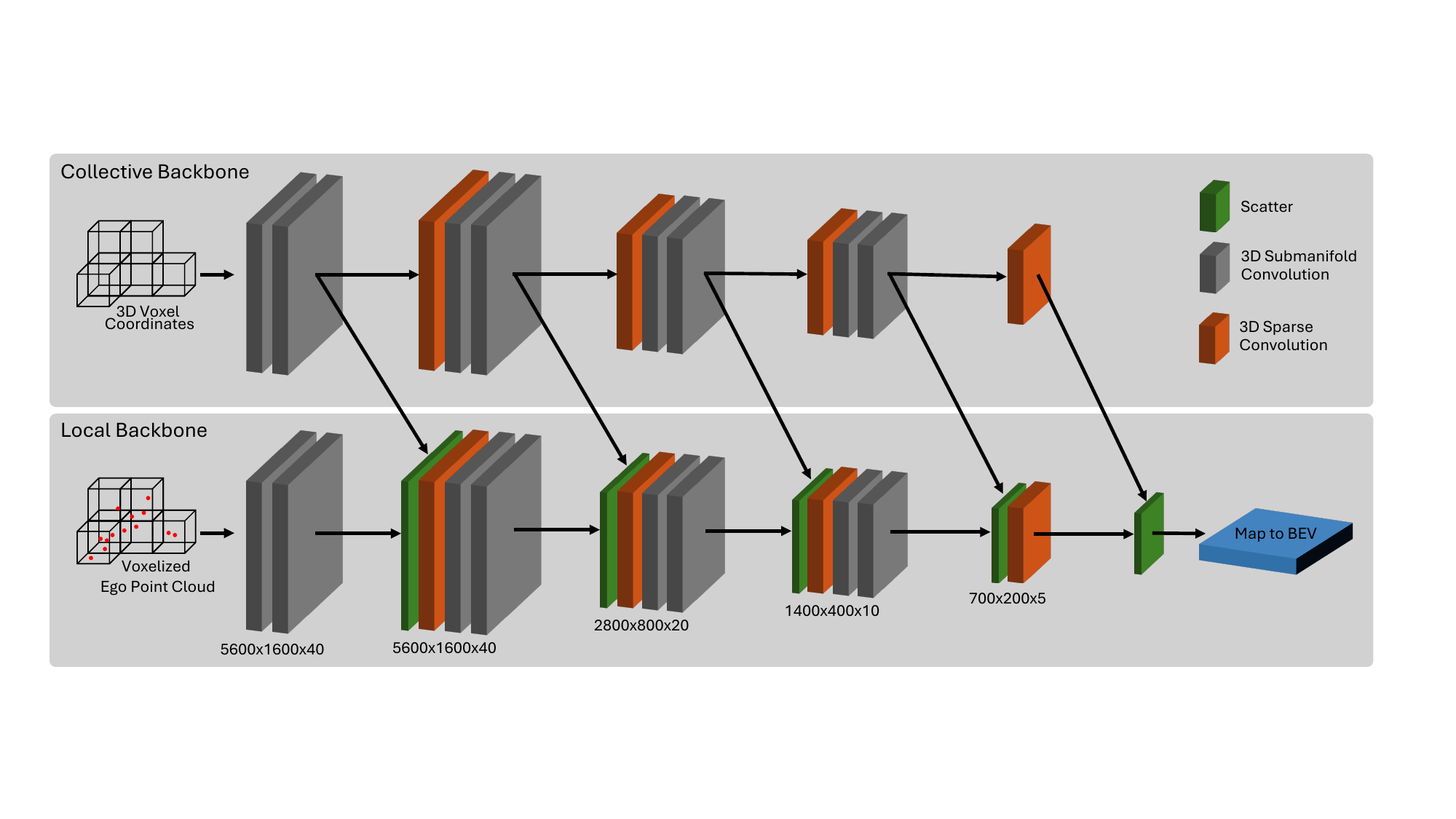}
    \caption{Architecture of S2S-Net. The collective backbone consists of one input streams which take sparse voxel grids as input. The local backbone takes the voxelized ego point cloud as input and shares the structure with the collective backbone. The outputs of the two backbones are fused using the scatter operation after each convolution block and then used as input for the subsequent layers in the local backbone.}
    \label{fig:architecture_overview}
\end{figure*}
We propose a lightweight and sensor-domain robust fusion architecture based on MR3D-Net \cite{mr3dnet}. In order to reduce the computational complexity and memory consumption, we omit the multiple resolution input streams from MR3D-Net and only use a single resolution input stream for the collectively shared sparse voxel grids. As shown in MR3D-Net, there is a trade-off between the level of detail and the data size of the sparse voxel grids. Since the change in detail might also affect the domain adaptation capabilities of S2S-Net, we evaluate S2S-Net with the high input resolutions for the collective backbone, as this is expected to be most prone to the Sensor2Sensor domain gap since it does not unify different sensor resolutions as much as the lower resolutions.
An overview of the proposed S2S-Net architecture is given in Figure \ref{fig:architecture_overview}.
We implemented S2S-Net into the OpenPCDet\cite{openpcdet2020} toolbox for LiDAR-based 3D object detection.

\subsection{Environment Representation}
In contrast to other existing CP methods S2S-Net utilizes sparse voxel grids as environment representation instead of raw sensor data, feature maps or object states. As proven in \cite{mr3dnet}, sparse voxel grids are an effective and compact environment representation for CP. The sparse voxel grids are built by partitioning the space of a point cloud into a uniform grid and only store the coordinates of the voxels containing points.
Sparse voxel grids have some crucial advantages over the other environment representations. Sparse voxel grids are much more compact than point clouds, achieving a data reduction of up to \SI{94}{\percent} while still preserving a high level of detail \cite{mr3dnet}. Furthermore, sparse voxel grids are a unified representation that can be fused with various models, without suffering from feature domain gaps as in intermediate fusion approaches. In S2S-Net the sparse voxel grids do not contain any features, only the coordinates of the voxel grids are exchanged to reduce the amount of transmitted data. As input to the model, the center points of the voxels are used as features, which can be calculated from the coordinates, voxel size and origin of the voxel grid. 

\subsection{Architecture}
In order to fuse the collective data with the local sensor data we leverage two input streams, the collective backbone and the local backbone. The collective backbone processes the collectively shared information whereas the local backbone processes the ego sensor data. Each input stream consists of four consecutive convolution blocks, where a convolution block consists of a sparse convolution layer followed by two sub-manifold convolutions. After each of these layers, a batchnorm and a relu activation function are applied. To reduce the spatial dimensions of the voxel grid, the sparse convolution layer may be strided. The collective input stream and the local input stream do not necessarily share the same resolution. If the input resolutions do not match we use different strides in the collective and local input stream to reduce the spatial resolution to a common resolution. In order to fuse the information between the different resolution streams, we use the scatter operation.

\subsubsection{Scatter Operation}
Due to the sparsity in the voxel grids a feature channel concatenation of two voxel grids is not possible since some voxels are present in only one of the two voxel grids and others are present in both voxel grids. Thus, a feature channel concatenation would cause an imbalance in the number of features per voxel, making it impossible to apply convolutions. To fuse the sparse voxel grids instead we apply the scatter operation, where on voxels that are present in both voxel grids, so called duplicate voxels, an element-wise $max$ function is applied, preserving the number of feature channels. The remaining voxels that only appear in either one of the voxel grids, so called single voxels, are unchanged. The final voxel grid is then just the union over the scattered duplicate voxels and the single voxels from both grids.

In S2S-Net the scatter operation is applied after each convolution block where the collective backbone and the local backbone share the same spatial resolution. The scattered sparse voxel grid is then used as input to the next convolution block in the local backbone, propagating the collective information towards the local backbone at different resolutions. This way, the input streams can learn from different inputs, enabling the neural network to extract features from multiple sources differently. After the four convolution blocks the result is processed by a final sparse convolution layer and then mapped into bird's eye view by concatenating the voxel features along the z-axis, resulting in a 2D feature map that is then used as input to the object detector.

\subsubsection{Implementation Details}
For the two input streams, we selected the voxel sizes to be $\SI{5}{\centi\meter} \times \SI{5}{\centi\meter} \times \SI{10}{\centi\meter}$. For training and evaluation, we used a maximum grid size of $\SI{280}{\meter} \times \SI{80}{\meter} \times \SI{4}{\meter}$, resulting in an input resolution of $\SI{5600}{} \times \SI{1600}{} \times \SI{40}{}$, voxels. In the local backbone and the collective backbone, the second, third, and fourth sparse convolutions use a stride of two to reduce spatial dimensions. The two input streams share the same number of output channels for all sparse convolution blocks, which are $[16, 32, 64, 64]$. The kernel size of all layers is $3\times3\times3$.

\begin{table*}[]
\small
\centering
\caption{Average Precision (AP) results at an IoU threshold of 0.7 on the SCOPE test set for the Car class. The best result for each Train/Test sensor combination is highlighted in bold.}
\label{tab:results}
\renewcommand{\arraystretch}{1.5}
\begin{tabular}{cccccc}
\multicolumn{1}{l}{}                          & \multicolumn{1}{l}{}                & \multicolumn{4}{c}{Sensor Test}                                   \\ \hline
\multicolumn{1}{c|}{Method}                   & \multicolumn{1}{c|}{Sensor Train}   & HDL64          & VLP32          & Blickfeld CUBE & Random Sensor  \\ \hline
\multicolumn{1}{c|}{}                         & \multicolumn{1}{c|}{HDL64}          & 63.61          & 56.38          & 19.07          & 51.14          \\
\multicolumn{1}{c|}{Attentive Fusion \cite{xu2022opv2v}}         & \multicolumn{1}{c|}{VLP32}          & \textbf{63.95} & \textbf{64.14} & 17.96          & 54.59          \\
\multicolumn{1}{c|}{}                         & \multicolumn{1}{c|}{Blickfeld CUBE} & 6.37           & 6.80           & 18.90          & 8.11           \\ \hline
\multicolumn{1}{c|}{}                         & \multicolumn{1}{c|}{HDL64}          & 64.23          & 56.76          & 16.23          & 49.45          \\
\multicolumn{1}{c|}{V2VAM \cite{li2023learning}}                    & \multicolumn{1}{c|}{VLP32}          & 62.29          & 61.36          & 15.02          & 52.23          \\
\multicolumn{1}{c|}{}                         & \multicolumn{1}{c|}{Blickfeld CUBE} & 8.92           & 8.85           & \textbf{21.72} & 9.40           \\ \hline
\multicolumn{1}{c|}{}                         & \multicolumn{1}{c|}{HDL64}          & 60.36          & 52.59          & 23.00          & 48.37          \\
\multicolumn{1}{c|}{Where2comm \cite{hu2022where2comm}}               & \multicolumn{1}{c|}{VLP32}          & 57.42          & 57.06          & 20.05          & 48.62          \\
\multicolumn{1}{c|}{}                         & \multicolumn{1}{c|}{Blickfeld CUBE} & 10.79          & 10.71          & 20.93          & 10.72          \\ \hline
\multicolumn{1}{c|}{}                         & \multicolumn{1}{c|}{HDL64}          & \textbf{66.61} & \textbf{68.96} & \textbf{61.09} & \textbf{67.36} \\
\multicolumn{1}{c|}{S2S-NET (Ours)}           & \multicolumn{1}{c|}{VLP32}          & 62.03          & 63.01          & \textbf{50.65} & \textbf{60.91} \\
\multicolumn{1}{c|}{}                         & \multicolumn{1}{c|}{Blickfeld CUBE} & \textbf{22.82} & \textbf{30.30} & 13.68          & \textbf{23.28}
\end{tabular}
\vspace*{4mm}
\end{table*}

\section{Evaluation}
\label{sec:eval}
\subsection{Dataset}
For the evaluation we use the SCOPE dataset~\cite{SCOPE-Dataset}, as discussed in section \ref{rel_work}. The SCOPE dataset is perfectly suited to evaluate domain adaptation capabilities in V2V collective perception due to the utilization of three different LiDAR sensors for each CAV. Furthermore, due to the diverse scenarios, various road users and varying environmental conditions the SCOPE dataset enables for an expressive evaluation of 3D object detection in CP. 

\subsection{Experiments}
To evaluate the generalization performance we train on each sensor individually, where both the ego sensor as well as the other CAV's sensors are the same. Then during testing, we keep the ego sensor the same as in the training stage and evaluate the model for each of the other CAV's sensors separately. The evaluation of the other CAV's sensor which is the same as the ego sensor, acts as a baseline, since this is the training domain, the other two sensors are then the unseen domains. Furthermore, since in a realistic scenario also the CAVs do not necessarily share the same sensors, we also evaluate a random assignment of the CAVs to the available sensors domains, which includes data from the training domain as well as from the two unseen sensor domains.
We do not evaluate the domain adaptation capabilities when changing the ego sensor in training and testing, since the ego sensor is always known and the model can be trained on this sensor domain, only the sensor domain of the other CAVs is unknown. 

We train S2S-Net on the SCOPE train split. We use S2S-Net as backbone together with PV-RCNN \cite{shi2023pv} as object detector. We compare S2S-Net with other state-of-the-art approaches for collective perception. We selected the sate-of-the-art methods Attentive Fusion \cite{xu2022opv2v}, V2VAM \cite{li2023learning} and Where2comm \cite{hu2022where2comm} due to their performance on other collective perception datasets and their ability to be trained with a high number of CAVs, since some state-of-the-art methods did not scale well with the number of CAVs and were therefore not trainable with up to 20 CAVs on a NVIDIA H100 GPU.  Even though S2S-Net supports multi-class object detection we only compare single class object detection for the "Car" class in order to get a fair comparison, since the other methods only support single-class object detection and are optimized for the detection of cars on other CP datasets. All methods were trained using a batch size of 6, for the other hyper-parameters the default values were used. Additionally, we also provide multi-class object detection results for S2S-Net to serve as a baseline for future works.

For evaluation, we report the Average Precision (AP) results on the SCOPE test split. As Intersection over Union (IoU) threshold we use \SI{0.7}{} for cars and vans and \SI{0.5}{} for pedestrians, (motor-)bikes and cyclists. We evaluate the detections within a range of $[-140, 140]$\,\si{\metre} in x-direction, $[-40, +40]$\,\si{\metre} in y-direction and $[-4, 1]$\,\si{\metre} in z-direction as this is the official evaluation range from SCOPE. For training and testing we do not limit the communication range between CAVs, i.e. all available information within the evaluation range are fused. For training, we filter all bounding boxes that do not contain at least one point from the ego sensor or one voxel from the shared voxel grids. In order to get a fair comparison of the different sensors used, we do not filter any bounding boxes during testing, even if they are fully occluded and there is no point or voxel inside. This is in contrast to most other collective perception benchmarks; however, filtering occluded bounding boxes would favor sensors with a lower resolution and FoV since e.g. blind spots would not impact the perception performance. This and the large evaluation range will cause the results to be relatively low compared to other collective perception benchmarks like OPV2V \cite{xu2022opv2v}.

\begin{table}[]
\centering
\caption{Multi-class object detection Average Precision on the SCOPE test set. For the classes Car and Van an IoU threshold of 0.7 was used, for Pedestrians, Cyclists and Motorbikes an IoU threshold of 0.5 was used.}
\label{tab:multi-class}
\renewcommand{\arraystretch}{1.5}
\resizebox{\linewidth}{!}{%
\begin{tabular}{@{}ccccc@{}}
\toprule
                                    &                                 &       & Sensor Test &                \\ \midrule
\multicolumn{1}{c|}{Sensor Train}   & \multicolumn{1}{c|}{Class}      & HDL64 & VLP32       & Blickfeld CUBE \\ \midrule

\multicolumn{1}{c|}{}               & \multicolumn{1}{c|}{Car}        & 68.38 &  67.79           &  58.43              \\
\multicolumn{1}{c|}{}               & \multicolumn{1}{c|}{Pedestrian} & 37.98 &  34.41           &  35.34              \\
\multicolumn{1}{c|}{HDL64}          & \multicolumn{1}{c|}{Cyclist}    & 26.27 &  25.27           &  15.12             \\
\multicolumn{1}{c|}{}               & \multicolumn{1}{c|}{Van}        & 70.83 &  68.53           &  62.25             \\
\multicolumn{1}{c|}{}               & \multicolumn{1}{c|}{Motorbike}  & 51.53 &  51.05           &  48.19              \\ \midrule

\multicolumn{1}{c|}{}               & \multicolumn{1}{c|}{Car}        & 64.72 & 65.26       & 53.98          \\
\multicolumn{1}{c|}{}               & \multicolumn{1}{c|}{Pedestrian} & 46.25 & 46.94       & 39.25          \\
\multicolumn{1}{c|}{VLP32}          & \multicolumn{1}{c|}{Cyclist}    & 24.55 & 24.10       & 13.40          \\
\multicolumn{1}{c|}{}               & \multicolumn{1}{c|}{Van}        & 67.75 & 76.32       & 63.63          \\
\multicolumn{1}{c|}{}               & \multicolumn{1}{c|}{Motorbike}  & 49.59 & 53.36       & 38.22          \\ \midrule

\multicolumn{1}{c|}{}               & \multicolumn{1}{c|}{Car}        & 37.53 &  45.76      & 21.82          \\
\multicolumn{1}{c|}{}               & \multicolumn{1}{c|}{Pedestrian} & 20.41 &  21.59      & 15.74          \\
\multicolumn{1}{c|}{Blickfeld CUBE} & \multicolumn{1}{c|}{Cyclist}    & 6.68  &  8.10       & 4.22           \\
\multicolumn{1}{c|}{}               & \multicolumn{1}{c|}{Van}        & 28.22 &  36.88      & 20.46          \\
\multicolumn{1}{c|}{}               & \multicolumn{1}{c|}{Motorbike}  & 24.06 &  26.12      & 17.96          \\ \bottomrule
\end{tabular}}
\vspace*{4mm}
\end{table}

\section{Results}
\label{sec:results}
An overview of the sensor domain adaptation results for S2S-Net and the other state-of-the art approaches is given in Tab. \ref{tab:results}. For the in-domain results of the HDL64 LiDAR sensor, S2S-Net achieved the best result with an AP of 66.61, followed by V2VAM and Attentive Fusion with APs of 64.23 and 63.61. For the VLP32 LiDAR sensor in-domain Attentive Fusion slightly outperformed S2S-Net by around 1\,p.p.. For the Blickfeld CUBE LiDAR in-domain, V2VAM performed the best with an AP of 21.72, followed by Where2comm and Attentive Fusion, S2S-Net performed the worst with an about 8\,p.p. lower AP compared to V2VAM. 
For the out-of-domain results trained on the HDL64 sensor and tested on the VLP32 sensor, S2S-Net clearly outperformed all other approaches with an AP of 12-16\,p.p. higher than the other approaches. Despite the domain gap between the HLD64 and the VLP32 being relatively small, since both sensors are \SI{360}{\degree} rotating sensors, all state-of-the-art methods clearly suffer from the domain gap induced by the lower resolution of the VLP32 sensor. Vice-versa, this cannot be observed, when trained on the VLP32 sensor and tested on the HDL64 sensor, all methods show approximately the same result as the VLP32 in-domain result. That there is a big domain gap between the Blickfeld CUBE sensor and the other two sensors can be clearly seen by the results of the state-of-the-art methods. While S2S-Net maintains a high AP of 61.09 when trained on the HDL64 sensor, the other methods drop drastically to an AP of 23.00, 19.07 and 16.23 for Where2comm, Attentive Fusion and V2VAM. For the training on the VLP32 sensor a similar behavior can be observed, however, here also the S2S-Net performance drops stronger than before. When training on the Blickfeld CUBE and testing on the other sensors this domain gap also drastically impacts the state-of-the-art methods. S2S-Net achieves an AP of 22.82 and 30.30 on the HDL64 and VLP32, while the other methods AP ranges from about 6 to 11 for both sensors. For the case where the CAVs sensors are selected randomly during testing, S2S-Net could also clearly outperforms the other methods with an increase of AP between 5.5-16 p.p. These results show the drastic impact of the Sensor2Sensor domain gap on current CP methods. Due to the unified environment representation of sparse voxel grids, S2S-Net maintains strong performance in unseen domains, where the other existing methods fail. These results also highlight the importance of the Sensor2Sensor domain gap for the development of future CP methods.
The multi-class object detection results for S2S-Net on the SCOPE dataset are given in Tab. \ref{tab:multi-class}. For the Car class the results are in accordance with the single class object detection results from Tab. \ref{tab:results}. 
The domain adaptation capabilities of S2S-Net for the other classes is similar to the car class with no huge performance drop due to the Sensor2Sensor domain gap. The best detection performance was achieved for the Van class with a slight increase compared to the Car class for the VLP32 and the HDL64, which was expected due to the larger size of vans. For the Blickfeld CUBE a slight decrease could be observed, which might arise from the small vertical FoV. For motorbikes and pedestrians the detection performance is about 10-15\,p.p. lower compared to cars, mostly due to their smaller size. For cyclists the detection performance is significantly lower compared to the other classes. Most likely this is caused by the fact, that the bicycle only has a small surface area, which results in only few lidar points on the bicycle, and the class similarity to both pedestrians and motorbikes.

\section{Conclusion and Future Work}
\label{sec:conclusion}
In this work, we presented S2S-Net, a sensor-domain robust fusion architecture using sparse voxel grids as a compact and unified environment representation for collective perception. This is the first work addressing the Sensor2Sensor domain gap in V2V collective perception.
This study showed, that all the evaluated state-of-the-art methods highly suffer from the Sensor2Sensor domain gap.
We demonstrated that S2S-Net is very robust to changes in the sensor domain and that it could significantly outperform all other evaluated methods. Especially when the sensor domains were highly different, S2S-Net achieved up to 44 percentage points higher average precision than the other methods. 
For future research, we will extend our evaluation to different voxel grid resolutions. Furthermore, we want to incorporate more fusion methods into our study to further characterize the sensor domain gap in V2V collective perception.
Moreover, we aim to conduct experiments incorporation infrastructural sensors to also study the domain gap arising from different placements of sensors. 









\bibliographystyle{IEEEtran}
\bibliography{literature}

\end{document}